\documentclass[runningheads]{llncs}

\usepackage{graphicx}
\usepackage{comment}
\usepackage{amsmath,amssymb}
\usepackage{color}
\usepackage{url}
\usepackage{hyperref}
\usepackage{algorithm}
\usepackage{algpseudocode}
\usepackage{xspace}

\newcommand*{\eg}{e.g.\@\xspace}
\newcommand*{\ie}{i.e.\@\xspace}

\newcommand*{\etc}{etc.\@\xspace}

\newcommand{\wrt}{w.r.t.\@\xspace} 
 
\newcommand{\etal}{et al.\@\xspace}

\newcommand{\arr}[2]{\begin{array}{#1} #2\end{array}}
\newcommand{\mat}[2]{\left[\!\!\arr{#1}{#2}\!\!\right]}
\newcommand{\V}[1]{\mbox{\boldmath$#1$}}        
\newcommand{\m}[1]{{\tt #1}}                    

\algnewcommand\algorithmicforeach{\textbf{for each}}
\algdef{S}[FOR]{ForEach}[1]{\algorithmicforeach\ #1\ \algorithmicdo}

\newif\ifreview
\reviewfalse

\ifreview
	\usepackage{lineno}

	\linenumbers
\fi

\begin{document}


\def\SubNumber{43}

\def\GCPRTrack{Main Track}

\title{D-InLoc++: Indoor Localization in Dynamic Environments}

\ifreview
\else
	\titlerunning{D-InLoc++: Indoor Localization in Dynamic Environments}

	\author{Martina Dubenova\orcidID{0000-0002-9936-292X} \and
Anna Zderadickova\orcidID{0000-0001-6661-3078} \and
	Ondrej Kafka\orcidID{0000-0003-4345-2309} \and
Tomas Pajdla\orcidID{0000-0001-6325-0072} \and
Michal Polic\orcidID{0000-0003-3993-337X}}
	
	\authorrunning{M. Dubenova et al.}
	
	\institute{CIIRC - Czech Institute of Informatics, Robotics and Cybernetics, Czech Technical University in Prague, Visual Recognition Group, Faculty of Electrical Engineering, Czech Technical University in Prague. 
	\email{dubenma1@fel.cvut.cz}}
\fi

\maketitle              

\begin{abstract}
Most state-of-the-art localization algorithms rely on robust relative pose estimation and geometry verification to obtain moving object agnostic camera poses in complex indoor environments. However, this approach is prone to mistakes if a scene contains repetitive structures, e.g., desks, tables, boxes, or moving people. We show that the movable objects incorporate non-negligible localization error and present a new straightforward method to predict the six-degree-of-freedom (6DoF) pose more robustly. We equipped the localization pipeline InLoc with real-time instance segmentation network YOLACT++. The masks of dynamic objects are employed in the relative pose estimation step and in the final sorting of camera pose proposal. At first, we filter out the matches laying on masks of the dynamic objects. Second, we skip the comparison of query and synthetic images on the area related to the moving object. This procedure leads to a more robust localization. Lastly, we describe and improve the mistakes caused by gradient-based comparison between synthetic and query images and publish a new pipeline for simulation of environments with movable objects from the Matterport scans. All the codes are available on \url{github.com/dubenma/D-InLocpp}. 

\end{abstract}
\section{Introduction}
Accurate camera pose is required for many applications, such as self-driving cars~\cite{badue2021self}, navigation of mobile robots~\cite{bonin2008visual}, or augmented reality~\cite{vidal2020analysis}. The global navigation satellite systems (GNSS) provide camera position within 4.9m for GPS~\cite{GPS_accuracy} and 4m for Galileo~\cite{galileo_accuracy} while camera rotation remains unknown. The accuracy of localization from Wi-Fi \cite{Wifi_localization}, or Bluetooth \cite{bluetooth_localization}, based on the signal strength varies a lot depending on the number and distance from signal broadcasters. Moreover, these methods provide the camera position only. The inertial measurement unit (IMU) allows camera pose tracking but cannot be employed for 6DoF localization. The last and most popular approach is visual localization, i.e., estimation of camera pose (i.e., position and orientation) given an RGB image. 

Visual localization~\cite{inloc,hloc,hyeon2021pose} is a challenging task since even small changes in camera pose lead to significant 3D mapping or navigation errors. In addition, the indoor environment is often repetitive \cite{torii2013visual}. Moreover, many plain walls, windows, or mirrors do not provide helpful information for localization. Another challenge is that the environment is only rarely static and often contains a lot of movable objects.

The main goal of this paper is to improve the visual localization so that it becomes more robust in dynamic environments, i.e., environments with moving objects. The motivation for this step is that moving objects often lead to significant changes in the scene, which make the localization algorithm work inaccurately. Also, dynamic environments are more common in the real world, making this improvement even more critical. We demonstrate a new method by improving the state-of-the-art localization pipeline InLoc~\cite{inloc}.

\section{Related Work}
\label{sec:related-work}
The recently published localization pipelines show remarkable performance in indoor spaces and urban environments~\cite{hyeon2021pose,hloc,sarlin2021back}. The recent state-of-the-art algorithms follow a common scheme:
\begin{itemize}
    \item \textbf{Image retrieval} step describes query image by feature vector and finds the most similar images in the database ~\cite{arandjelovic2016netvlad,radenovic2018fine,revaud2019learning,hyeon2020kr}.
    \item \textbf{Relative pose} step finds local correspondences~\cite{lowe2004distinctive,detone2018superpoint,dusmanu2019d2,revaud2019r2d2,sun2021loftr,laguna2022key} between the query image and the $k$ most similar images selected in the image retrieval step. Furthermore, the correspondences are verified by a robust model estimator~\cite{yi2018learning,sun2020acne,chum2003locally}, \cite{chum2005two,barath2018graph,barath2020magsac++,tong2021deep,sarlin2020superglue} using a suitable relative pose solver~\cite{hartley2003multiple,kukelova2008polynomial,kukelova2015efficient}.
    \item \textbf{Absolute pose} step constructs 2D-3D correspondences from 2D-2D correspondences found in previous step. The correspondences are calculated for $t$ database images with the most inliers. Further, an absolute pose solver~\cite{larsson2017making} in a robust model estimator provides a camera pose proposals. 
    \item\textbf{Pose correction} step optimizes~\cite{hyeon2021pose,lindenberger2021pixel,triggs1999bundle} the camera pose proposals.
    \item \textbf{Pose selection} step evaluates the camera pose proposals by photometric consistency and selects the most fitting candidate~\cite{pyrender,arandjelovic2012three,liu2010sift}.
\end{itemize}
The last two steps are not included in run-time methods, \eg,~\cite{hloc}. Recently published localization algorithms focus on improving specific modules while the general scheme remains the same. 

The images are usually described by NetVLAD~\cite{arandjelovic2016netvlad} but several extensions, \eg, GeM~\cite{radenovic2018fine}, AP-GeM~\cite{revaud2019learning} and i-GeM~\cite{hyeon2020kr} were published. Relative pose estimation has experienced the largest evolution in recent years~\cite{jin2021image}. 

Key point detectors evolved from handcrafted, \eg, SIFT~\cite{lowe2004distinctive}, to the learned ones, \eg, SuperPoint~\cite{detone2018superpoint}, D2-Net~\cite{dusmanu2019d2}, R2D2~\cite{revaud2019r2d2}, LoFTR~\cite{sun2021loftr} or a Key.Net~\cite{laguna2022key} paired with HardNet or SOSNet descriptors. 
The matching can start by Context Networks (CNe)~\cite{yi2018learning,sun2020acne} to pre-filter the outliers followed by RANSAC based (Lo-RANCAS~\cite{chum2003locally}, DEGNSAC~\cite{chum2005two}, GC-RANSAC~\cite{barath2018graph}, MAGSASC++~\cite{barath2020magsac++} or Deep-MAGSAC~\cite{tong2021deep}) or neural-network based (\eg, SuperGlue~\cite{sarlin2020superglue}) matches verifier. There are also the end-to-end detectors + matchers, \eg, Sparse-NCNet~\cite{rocco2020efficient} or Patch2Pix~\cite{zhou2021patch2pix}. Moreover, a list of relative pose solvers employed in the scope of RANSAC-based algorithm, \eg, H4P~\cite{hartley2003multiple}, E5P~\cite{kukelova2008polynomial}, F7P~\cite{hartley2003multiple} or F10e~\cite{kukelova2015efficient}, is available. Each combination of listed relative pose algorithms lead to different camera pose accuracy and robustness. The most common among learned key-point detectors is the SuperPoint~\cite{detone2018superpoint} and the matching algorithm is the SuperGlue~\cite{sarlin2020superglue}. As far as we know, the latest approaches for detection and matching of keypoints~\cite{jin2021image,laguna2022key,tong2021deep} were not published in a localization pipeline yet. 

The absolute pose step estimates the camera pose by robust model estimator (usually the Lo-RANSAC~\cite{chum2003locally}). The constraints on a camera define the minimal solver used, \eg, P3P, P4Pf~\cite{larsson2017making} or PnP~\cite{hartley2003multiple}. 

The pose correction step optimizes previously estimated camera poses. The optimization requires additional time to run and improves the results mainly in sparsely mapped indoor environments with texture-less areas, \eg, in the InLoc~\cite{inloc} dataset. Therefore, it is employed mainly for offline indoor localization. The Pixel-Perfect-SfM~\cite{lindenberger2021pixel} adjusts keypoints by featuremetric refinement followed by the featuremetric Bundle Adjustment (BA)~\cite{triggs1999bundle}. The PCLoc~\cite{hyeon2021pose} generates "artificial images" close to the camera pose proposal composed of reprojected feature points. Further, the relative and absolute pose step runs again to calculate camera pose from generated artificial image. As far as we know, the combination of both mentioned methods was not published yet.  

The last step of this standard scheme is the photometric verification of the camera pose proposals. The online algorithms, \eg, the Hierarchical Localization~\cite{hloc}, do select the camera pose with the most inliers after the absolute pose estimation step. The photometric verification requires a 3D model of the environment and appears beneficial in indoor spaces~\cite{inloc}. The idea is to render synthetic images for the $t$ camera pose proposals and compare them by pixel-wise patch descriptors, \eg, DenseRootSIFT~\cite{arandjelovic2012three,liu2010sift}, with the query image. The rendering can be done either by standard software as Pyrender~\cite{pyrender} (implemented on InLoc), or Neural Rerendering in the Wild~\cite{martin2021nerf}, or the NeRF in the Wild~\cite{martin2021nerf} to obtain realistically looking synthetic images. As far as we know, the NeRF-based approaches were not published in a localization pipeline yet. 

As mentioned in the previous paragraph, the map format plays an important role. The photometric verification cannot run if the 3D model of the environment is unknown. The paper Torii \etal~\cite{torii201524} shows that more database images in the map lead to more accurate and robust localization. In the case of dynamic environment, the map can be composed by \cite{runz2018maskfusion,palazzolo2019refusion}.

The last direction of the research focuses on localization from a sequence of images. Having a set of pictures with known poses in the local coordinate system and the 2D-3D correspondences, we can utilize the generalized absolute pose solver, \eg, gp3p or gp4ps~\cite{kukelova2016efficient}, to get the global pose of the sequence of images. The 2D keypoints are converted to 3D rays and aligned with 3D points from the map by Euclidean transformation. This approach was published in Stenborg \etal~\cite{stenborg2020using}. \\

Generally, there exists a number of extensions~\cite{jin2021image,tong2021deep,kukelova2015efficient,martin2021nerf,lindenberger2021pixel} that can be used to improve individual modules of the state-of-the-art localization~\cite{localization_benchmark}. However, we would like to avoid mechanical replacement of particular methods and open new unexplored yet important topics. As far as we know, the scenes with movable objects were not addressed in detail yet. Our contributions are:
\begin{itemize}
    \item the evaluation of localization accuracy in the environment with movable objects   
    \item implementation of D-InLoc++ that is robust against the mismatches caused by movable objects
    \item new approach to comparing the synthetic and query images
    \item automated pipeline for testing the localization w.r.t. movable objects
\end{itemize}
The rest of the paper is organized as follows. The following section describes the moving object filtering during the localization process. Further on, we show the challenges related to the usage of DenseRootSIFT and propose a new solution for comparing synthetic and query images. The fifth section describes a new pipeline for generating synthetic datasets with movable objects. 
The last section evaluates dynamic object filtering on the published datasets.  

\section{Dynamic Object Filtering}
We propose a simple yet effective approach to dealing with movable objects. The pseudocode of the algorithm is shown in Algorithm~\ref{alg:localization-pseudocode}. Let us assume that we have a static map realized as in the InLoc dataset~\cite{inloc}. The map consists of $N$ RGB-D images $\mathcal{I} = \{I_1, \dots, I_N\}$ with known camera poses $\mathcal{P}_i = \{\m{R}_i, \V{t}_i\}$, \ie, the rotation $\m{R}_i \in \mathbb{R}^{3 \times 3}$ and translation $\V{t}_i \in \mathbb{R}^{3}$. Without loss of generality, we assume no radial distortion and images are captured by the same camera. The camera intrinsic parameters are the focal length $f \in \mathbb{R}$ and principal point $p \in \mathbb{R}^2$. The depth images in the map are converted into mesh $\mathcal{M}_{\mathcal{I}}$ by the Truncated Signed Distance Function (TSDF) algorithm. The set of $M$ query images is denoted by $\mathcal{Q} = \{Q_1, \dots, Q_M\}$.
If the same camera does not capture the query images and the database images in the map, the resolution is adjusted in advance. 
Let us assume that the query images $\mathcal{Q}$ already match the resolution of images $\mathcal{I}$ in the database. \\

At first we add a preprocessing step, \ie, we extend the map with the correspondences between the pairs of database images. The database images are described by feature vectors (\eg, by NetVLAD) and keypoints (\eg, by SuperPoint). For the top $k$ closest database images (computed as dot product between feature vectors) are found tentative matches (\eg, by SuperGlue). Next, we verify the tentative matches by relative pose constrains extracted from known camera poses. 

\begin{algorithm}[H]
\caption{The pseudo-code of D-InLoc++}
\label{alg:localization-pseudocode}
    \begin{algorithmic}[1] 
    \State \textbf{Inputs:} set of query images $\mathcal{Q}$; map: RGB-D images $\mathcal{I}$ and camera poses $\mathcal{P}_{\mathcal{I}}$; epipolar error threshold $\mathcal{T}$, min. mask size $\gamma$, moving object criterion threshold $\delta$
    \State \textbf{Outputs:} camera poses $\mathcal{P}_{\mathcal{Q}}$ for query images $\mathcal{Q}$;
    \State ~
    \State \textbf{Pre-process the map (offline):}
    \State Mesh $\mathcal{M}_{\mathcal{I}} \gets$ TSDF($\mathcal{I}$, $\mathcal{P}_{\mathcal{I}}$);
    \State Keypoints $\V{u}^{(\mathcal{I})}$, Visibility ids set $\mathcal{Y}_{\mathcal{I}}$, Points $\V{\mathcal{X}}_{\mathcal{I}}$ $\gets$ sparse\_reconstruction($\mathcal{I}$,$\mathcal{P}_{\mathcal{I}}$)
    \label{line:sparse_reconstruction}
    \State Masks $\V{\mathcal{S}}_{\mathcal{I}}$, $\V{\mathcal{D}}_{\mathcal{I}}$, $\V{\mathcal{U}}_{\mathcal{I}}$ $\gets$ instance\_segmentation($\mathcal{I}$)
    \State Masks $\V{\mathcal{S}}_{\mathcal{I}}$, $\V{\mathcal{U}}_{\mathcal{I}}$ $\gets$ reassign\_small\_masks($\gamma$, $\V{\mathcal{S}}_{\mathcal{I}}$, $\V{\mathcal{U}}_{\mathcal{I}}$)
    \State ~
    \State \textbf{Localize query images (online):}
    \State $\mathcal{P}_{\mathcal{Q}} \gets$ [] 
    \ForEach {$Q_j \in \mathcal{Q}$}
        \State Masks $\mathcal{S}_{Q_j}$, $\mathcal{D}_{Q_j}$, $\mathcal{U}_{Q_j}$ $\gets$ instance\_segmentation($Q_j$)
        \State Masks $\mathcal{S}_{Q_j}$, $\mathcal{U}_{Q_j}$ $\gets$ reassign\_small\_masks($\gamma$, $\mathcal{S}_{Q_j}$, $\mathcal{U}_{Q_j}$)
        \State $\V{u}^{(Q_j)} \gets$ compute\_keypoints($Q_j$)
        \State $\{T_1, \dots, T_k\} \in \mathcal{I} \gets$ find\_closest\_images($k$, $Q_j$, $\mathcal{I}$)
        \State $P_{Q_{j}} \gets$ []
        \State $L_{Q_{j}} \gets$ []
        \ForEach {$T \in \{T_1, \dots, T_k\}$}
            \State $\mathcal{Y}_{Q_j, T} \gets$ find\_correspondences\_2D3D(compute\_matches($\V{u}^{(Q_j)}$, $\V{u}^{(T)}$, $\mathcal{D}_{(T_i)}$, $\mathcal{D}_{(Q_i)}$), $\mathcal{Y}_{\mathcal{I}}$)
            \State $\beta_T \gets$ calculate\_moving\_object\_criteria($(\mathcal{Y}_{Q_j, T} \cup \mathcal{Y}_{\mathcal{I}})$, $\V{u}^{(Q_j)}$, $\mathcal{U}_T$)
            \State $\beta_{Q_j} \gets$ calculate\_moving\_object\_criteria($(\mathcal{Y}_{Q_j, T} \cup \mathcal{Y}_{\mathcal{I}})$, $\V{u}^{(Q_j)}$, $\mathcal{U}_{Q_j}$)
            \State $\mathcal{S}_T^*$, $\mathcal{D}_T^* \gets$ reassign\_unknown\_masks($\mathcal{S}_T$, $\mathcal{D}_T$, $\mathcal{U}_T$, $(\beta_T < \delta)$)
            \State $\mathcal{S}_{Q_j}^*$, $\mathcal{D}_{Q_j}^* \gets$ reassign\_unknown\_masks($\mathcal{S}_{Q_j}$, $\mathcal{D}_{Q_j}$, $\mathcal{U}_{Q_j}$, $(\beta_{Q_j} < \delta)$)
            \State $\mathcal{Y}^*_{Q_j, T} \gets$ filter\_moving\_objects\_matches($\mathcal{Y}_{Q_j, T}$, $\mathcal{D}_T^*$, $\mathcal{D}_{Q_j}^*$)
            \State $\mathcal{P}_{Q_{j}, T},\,\mathcal{Y}^+_{Q_j, T} \gets $ estimate\_absolute\_pose($(\V{u}^{\mathcal{I}} \cup \V{u}^{(Q_j)})$, $\V{\mathcal{X}}_{\mathcal{I}}$,  $\mathcal{Y}^*_{Q_j, T}$)
            \State $L_{Q_{j}} \gets$ [$L_{Q_{j}}$, count\_number\_of\_inliers($\mathcal{Y}^+_{Q_j, T}$) ]
            \State $P_{Q_{j}} \gets$ [$P_{Q_{j}}$, $\mathcal{P}_{Q_{j}, T}$]
        \EndFor
        \State $\bar{\mathcal{P}}_{Q_j} \gets $ sort\_poses\_descending($L_{Q_{j}}$, $\mathcal{P}_{Q_{j}}$)
        \State $\bar{L}_{Q_{j}} \gets$ []
        \ForEach {$p \in \{1, \dots, l\}$}
            \State $\bar{Q}_{j, p} \gets$ render\_image($\bar{\mathcal{P}}_{Q, p}$, $\mathcal{M}_{\mathcal{I}}$)
            \State $\bar{L}_{Q_{j}} \gets$ [$\bar{L}_{Q_{j}}$, compare\_synth\_and\_query\_image( $\bar{Q}_{j, p}$, $Q_j$)]
        \EndFor
        \State $\tilde{\mathcal{P}}_{Q_j} \gets $ sort\_poses\_ascending($\bar{L}_{Q_{j}}$, $\bar{\mathcal{P}}_{Q_j}$)
        \State $\mathcal{P}_{\mathcal{Q}} \gets$ [$\mathcal{P}_{\mathcal{Q}}$, $\tilde{\mathcal{P}}_{Q_j, 1}$] 
    \EndFor
    \State \Return $\mathcal{P}_{\mathcal{Q}}$;
    \end{algorithmic}
\end{algorithm}

\noindent Let us define the calibration matrix $\m{K}$ and camera center $\V{C}_i$ as 
\begin{equation}
    \m{K} = \mat{ccc}{f & 0 & p_1\\ 0 & f & p_2\\ 0 & 0 & 1} \quad \quad and \quad \V{C}_i = - \m{R}_i^{\top} \V{t} .
\end{equation}
We assume the Fundamental matrix $\m{F}$ to equal
\begin{equation}
    \m{F} = \m{K}^{-\top} \m{R}_2 [\V{C}_2 - \V{C}_1]_{\times} \m{R}_1^{\top} \m{K}^{-1} ,
\end{equation}
and epipolar constrains for keypoint $\mat{ccc}{u^{(i)}_{1} & u^{(i)}_{2} & 1}^{\top}$ in the $i$-th and $\mat{ccc}{u^{(j)}_{1} & u^{(j)}_{2} & 1}^{\top}$ in the $j$-th image:
\begin{equation}
    \mat{ccc}{u^{(j)}_{1} & u^{(j)}_{2} & 1} \m{F} \mat{c}{u^{(i)}_{1} \\ u^{(i)}_{2} \\ 1} < \mathcal{T} .
    \label{eqn:relative-pose-constrain}
\end{equation}
The value $\mathcal{T}$ represents the threshold. If the equation~\ref{eqn:relative-pose-constrain} holds, the correspondence $\{\V{u}^{(i)}, \V{u}^{(j)}\}$ is assumed to be correct (\ie, inlier) and stored in the map. All keypoints detected on database images are called $\V{u}^{(\mathcal{I})}$. The map is further extended with points $\V{\mathcal{X}}_{\mathcal{I}}$ and the index set $(i,j) \in \mathcal{Y}_{\mathcal{I}}$ storing the information that $j$-th point $\V{X}_j \in \mathbb{R}^3$ is visible in $i$-th image $I_i$. This step is done by the function sparse\_reconstruction($\mathcal{I}$,$\mathcal{P}_{\mathcal{I}}$) on line~\ref{line:sparse_reconstruction} in the pseudocode of Algorithm~\ref{alg:localization-pseudocode}.

Further, in the case of moving objects in the map, we run the instance segmentation on database images $\mathcal{I}$ to have the object masks in advance. This method separates the background from possibly movable instances of the objects. We assume three classes of objects, the static one $\mathcal{C}_S$, dynamic (moving) one $\mathcal{C}_D$ and movable (unknown) $\mathcal{C}_U$. Related masks in the image $i$ are called $\mathcal{S}_i$, $\mathcal{D}_i$, and $\mathcal{U}_i$. The background is an example of static class $\mathcal{C}_S$. People and cars are an example of $\mathcal{C}_D$ and the rest of objects where we cannot decide in advance is in $\mathcal{C}_U$.\\

The online localization process starts with segmentation of the images $\mathcal{Q}$ by a real-time instance segmentation (\eg, the YOLACT++). We assume the same classes for the objects as for the database images. The masks for single query image $Q_j$ are called $\mathcal{S}_{Q_j}$ for static, $\mathcal{D}_{Q_j}$ for dynamic, and $\mathcal{U}_{Q_j}$ for movable objects. The image retrieval module extracts the feature vectors and finds the $k$ most similar images $\{T_1, \dots, T_k\}$ in the database, function find\_closest\_images($k$, $Q_j$, $\mathcal{I}$). Next, the algorithm calculates keypoints $\V{u}^{(Q_j)}$, matches, and 2D-3D correspondences $\mathcal{Y}_{Q_j, T}$ between $Q_j$ and the database image $T$. All the keypoints laying inside $\mathcal{D}_{T}$ and $\mathcal{D}_{Q_j}$ masks are not assumed for computation of matches. Next, we propose a simple criterion to decide which image areas out of $\mathcal{U}_T$ and $\mathcal{U}_{Q_j}$ will be utilised further. At first, we check the instance segmentation masks and all that are small enough, \ie, smaller than $\gamma$, are reassigned to $\mathcal{S}_T$ or $\mathcal{S}_{Q_j}$. For the rest of object masks in $\mathcal{U}_T$ and $\mathcal{U}_{Q_j}$ we calculate 
\begin{equation}
    \beta_{T_s} = g(u_{T_s}) - num\_px(u_{T_s})
    \label{eqn:mask-sel-criterion}
\end{equation}
where $g$ gives the number of observations with tracks of length $\geq 3$ and a correspondence in $Q_j$ that are laying in $s$-th object mask $u_{T_s} \in \mathcal{U}_T$. The vector for all the masks is called $\beta_T$ for $T$ image and $\beta_{Q_j}$ for $Q_j$ image. In other words, we count the number of observations that have our query image in the track, and the track has a length larger than three, \ie, the point in 3D is visible in more than two views. Such tracks should appear if the object is static. Large masks should contain more tracks fulfilling this condition. The $num\_px$ function counts the number of pixels in some object mask, \eg, $u_{T_s}$. All the object masks with $\beta_{T} > \delta$ are reassigned to $\mathcal{S}_T$ and $\beta_{Q_j} > \delta$ to $\mathcal{S}_{Q_j}$. Remaining masks are reassigned to $\mathcal{D}_T$ or $\mathcal{D}_{Q_j}$. This step creates masks $\mathcal{S}_T^*$, $\mathcal{S}_{Q_j}^*$ for each pair of query-database images. Only the keypoints and matches found in $\mathcal{S}_T^*$, $\mathcal{S}_{Q_j}^*$ are utilised in the next absolute pose estimation module. 

Further, the algorithm follows the standard localization scheme, see Section~\ref{sec:related-work}. The absolute pose module estimates the camera pose in the global coordinate system. The pose correction module is skipped in our reference algorithm InLoc~\cite{inloc}, and we do not assume it either. 
The last module of this scheme is related to pose selection out of the proposals, and we discuss it in the following section.

\section{Image Selection Step}
This step of the localization process selects the most appropriate pose among all the proposals and, therefore, directly influences the results of the rest of the pipeline. 
The standard approach to comparing the synthetic and real images starts with applying a path descriptor (\eg, the DenseRootSIFT) to all the patches in both images. This provides a feature vector that describes the surrounding area of each pixel. Then a median of the norms of related feature vector differences is taken. Smaller values indicate greater similarity. This approach is used in many algorithms, such as~\cite{inloc,hyeon2021pose}, for photometric comparison of a pair of images.  

\begin{figure}[tb]
\includegraphics[width=0.32\textwidth]{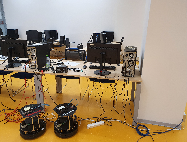}
\includegraphics[width=0.32\textwidth]{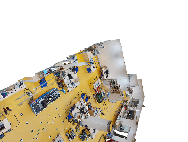}
\includegraphics[width=0.32\textwidth]{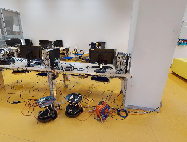}
\caption{The images from left to right correspond to: the query image, the first and the second most similar synthetic images sorted by the DenseRootSIFT criterion. It can be seen that a completely wrong image has a better score because the gradient of the background is almost identical to the gradient of the wall and pillar.} 
\label{fig:wrong-camera-comparison}

\end{figure}

However, this approach leads to several issues, as can be seen in Figure~\ref{fig:wrong-camera-comparison}. The first issue is that the gradient-based comparison of images does not take into account the colors. For example, the flat white walls have almost zero error \wrt the background of the synthetic image. This is caused by lack of gradient on both areas. If the indoor space consists of textureless corridors or rooms, the poses looking outside the mapped area have a small error and may be selected. Second, this approach does not consider movable objects and assumes the same weight for all the pixels. An example of such objects can be people, chairs, doors, and other equipment that is usually not static \wrt the walls, pillars, \etc.

Because the starting position for the image selection step is several images, for which similarity is to be pair-wisely compared, the first considered approach was to use some already published learnable methods for relative pose estimation. Various approaches also utilizing convolutional neural networks (CNN) for this task were examined in~\cite{shavit2019introduction}. Remarkable results regarding this issue were achieved in~\cite{chen2021wide}. However, in our case, we found it too narrowly focused on pose estimation. So for our task, we opted for a more general approach similar to the one adopted in~\cite{rpCNN}.
Therefore, the foundation of our method is a CNN used as an image encoder, EfficientNet~\cite{tan2019efficientnet}. This part of our method is a representational part, \ie, it outputs a vector representation of what is in the images. This part is then followed by a regression part - a fully connected layer.

To be more specific, the algorithm itself works as follows. Input is a set of the top $k$ proposed images $\mathcal{I}_k \subseteq \mathcal{I}$ and query image $Q_i \in \mathcal{Q} $.

Inputs to our network are then all $k$ pairs of images $( Q_i, I_j)$, where $I_j \in \mathcal{I}_k$. If rescaling is needed, both $Q_i$ and all $I_j$ are reshaped to fit the CNN input dimensions.
Then instance segmentation masks and masks of the moving objects in query image allow to select compared areas, by setting the pixels to white. 

These pairs are then sequentially processed by the representational part, producing feature vectors $\mathbf{f}_i, \mathbf{f}_j \in \mathbb{R}^{1280\times l_1 \times l_2}$. These are then fed into the regression part and output score $s \in \mathbb{R}$. 

The value of $s$ was trained to represent a measure of similarity of the images. The ground truth values (targets) from the training dataset were computed as $min(1, (\theta + 10 t)/50)$, the result of which is a value for each training pair ranging in the interval from 0 to 1. The formula provides a value of 1 for everything that exceeds the threshold of similarity, that is  $\theta + 10 t = 50$, where $\theta$ is the angular difference between the two images and $t$ distance between camera positions $\mathbf{t_i}, \mathbf{t_j} \in \mathbb{R}^3$ computed as
\begin{equation}
    \theta = \mathrm{arccos}\left(\frac{1}{2}(\mathrm{trace}(R) - 1)\right), \; t = |\mathbf{t_i} -\mathbf{t_j}|,
\end{equation}

where $R \in \mathbb{R}^{3 \times 3}$ is a rotation matrix representing rotation between cameras corresponding to $Q_i$ and $I_j$, and $\mathbf{t_i}, \mathbf{t_j} \in \mathbb{R}^{3}$ are vectors of camera positions. The parameter 10 in the mentioned formula was empirically set to balance the range of values $\theta $ and $t$.

Finally, the comparison of our CNN-based approach (combined with DenseRootSIFT - the CNN serves as guidance for DSIFT as to which images to consider) with other methods is in the Figure~\ref{fig:quantitative-eval}. These results were obtained on 180 query images and show the percentage of the cases when the chosen image pose was within the thresholds shown in the figure.

\section{Dynamic Dataset Generation}
Most of the current localization datasets~\cite{localizationDatasets} include entities such as people, doors, and chairs that are seldom static. However, the masks of stationary and moving objects are missing because manual labeling is time-consuming. We propose a new pipeline that takes as input standard mapping technology with additional models of movable objects. Our method generates camera poses and the RGB-D images with masks of movable objects. 

We chose the services provided by Matterport~\cite{matterport} as the scanning technology of indoor spaces. To create the 3D model of the environment, the customer can use a user-friendly application and a reasonably expensive scanner. The scanner captures 360 degrees RGB-D images aligned to a single coordinate system. However, these scans are not directly available as the RGB-D panoramic images. We provide the script for downloading camera poses from the Matterport API endpoint. Further, Matterport API provides six square pictures of the panoramic image projected onto the shape of a cube. These images follow the global coordinate system, and we projected them back into panoramic images by cube2sphere~\cite{cube2sphere} library. Each panoramic image is projected into 36 RGB perspective images.
These images are called cutouts~\cite{inloc}. The related depth images are obtained by rendering the mesh in the AI Habitat~\cite{habitat} simulator. 

In this way, we acquire a dataset representing a scene without any moving objects.
The simplest way to add moving objects to queries is by adding artificial textures of the moving objects in the query images. 
 
To have consistent data for the localization of RGB-D image sequences, we propose to add the movable objects directly into the 3D model of the environment to preserve the geometry. AI Habitat renders the depth, semantic masks, and RGB images. However, the RGB cutouts channels have higher quality than rendered images. Therefore, the script replaces any image data outside the masks of movable objects with original cutouts projected from panoramic images. An example of such an image can be seen in Figure~\ref{fig:dynamic-query}. 

Published codes simplify the creation of datasets with movable objects and speed up further development.

\section{Experimental Evaluation}
This section compares the original localization pipeline and the adjusted method by robust filtering of movable objects. A new dynamic localization dataset generator produced all the data utilized here.  

\begin{figure}[tb]
\includegraphics[width=\textwidth]{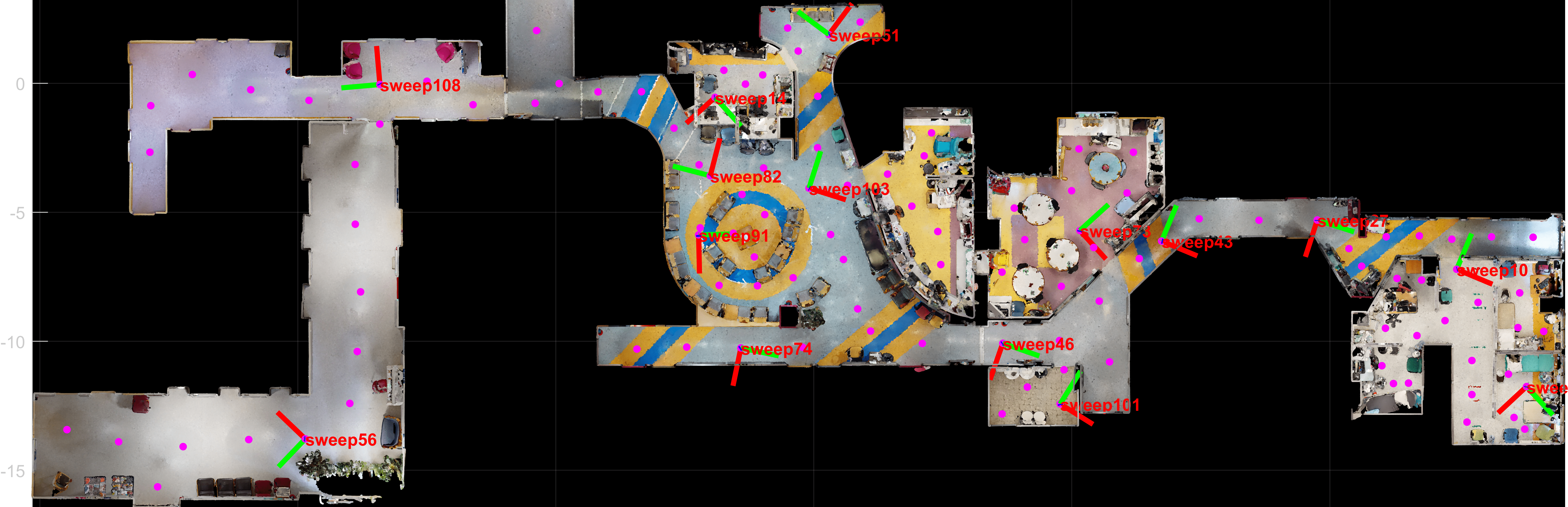}
\caption{The visualization of the Hospital space. Purple dots show the positions of Matterport scans that are in Matterport terminology referred to as sweeps. The sweeps that are used to generate query images  are highlighted by showing their coordinate systems.} \label{fig:hospital-space}
\end{figure}

\subsection{Composition of Test Datasets}
\label{sec:composition-of-test-dataset}
As far as we know, datasets with ground truth semantic masks for movable and static objects were not published yet~\cite{localization_benchmark}. We provide such a dataset. We recorded a real environment, \ie, the Hospital space. An example of reconstructed space is in Figure~\ref{fig:hospital-space}. For recording, we used the device the Matterport Pro 3D Camera~\cite{matterportcamera}. We aimed to have around 1.5 meters between the camera locations and cover the whole space during the capturing. The Hospital space contains 116 scans. It resulted in 696 images of skyboxes, \ie, images projected into cube faces, provided by Matterport API. These images have a resolution of 2048x2048 pixels. We used them to compose 116 panoramic images with a resolution of 8192x4096 pixels. Next, we downloaded the dense reconstruction model. To generate the cutouts, we adopted the same strategy as presented in InLoc~\cite{inloc}, \ie, generated 36 images out of one panoramic image using a sampling of 30 degrees in yaw and ±30 degrees in pitch angles. This gave us 4176 cutouts in total. To project the panoramic images into the perspective ones, we have chosen the HoloLens photometric camera intrinsic parameters, \ie, the focal length $f = 1034$, principal point $p = [672, 378]$ and images resolution $1344 \times 756$ pixels. Anyone can adjust these parameters in the dataset generation pipeline. 

Further, we randomly added objects inside the reconstructed area. We manually checked if the object collided with any other meshes and moved such objects to a new location. The rendering of the mesh with moving objects is implemented in AI Habitat and takes about 0.35 seconds/image on a personal laptop. AI Habitat requires the input in the .glb format. The script employs obj2gltf~\cite{obj2gltf} library for the conversion of all the models (\ie, objects, and the environment). The AI Habitat generates the synthetic images and masks for static and movable objects in a way that they correspond to each cutout. In the case of dynamic datasets, the rendered texture of moving objects replaces related areas in the cutouts of query images. The resulting image can be seen in Figure~\ref{fig:dynamic-query}. The database images do not contain any moving objects. 

Generally, we created two datasets: static and dynamic. The dataset without any dynamic objects is called a static dataset. The dynamic dataset has the objects placed on the ground with the average area of the objects masking 20.63$\%$ of the whole query image for ground truth masks and 13.09$\%$ from proposed instance segmentation. We used 15 of the Matterport scans to generate query images that all face only the horizontal direction to simulate a camera on a robot, \ie, 180 queries for the Hospital scene. The rest of the images is utilized for creating the localization map. 

\subsection{Evaluation}
The main part of this paper is to show the influence of movable objects appearing in the localization dataset. We compared four main scenarios, \ie, running 
\begin{enumerate}
    \item original Inloc,
    \item improvement with neural network (InLoc++),
    \item improvement with filtering of matches (D-InLoc),
    \item improvement with combination of 2) and 3) (D-InLoc++).
\end{enumerate}

\subsubsection{Ablation Study}
\begin{figure}[h]
\includegraphics[width=0.49\textwidth]{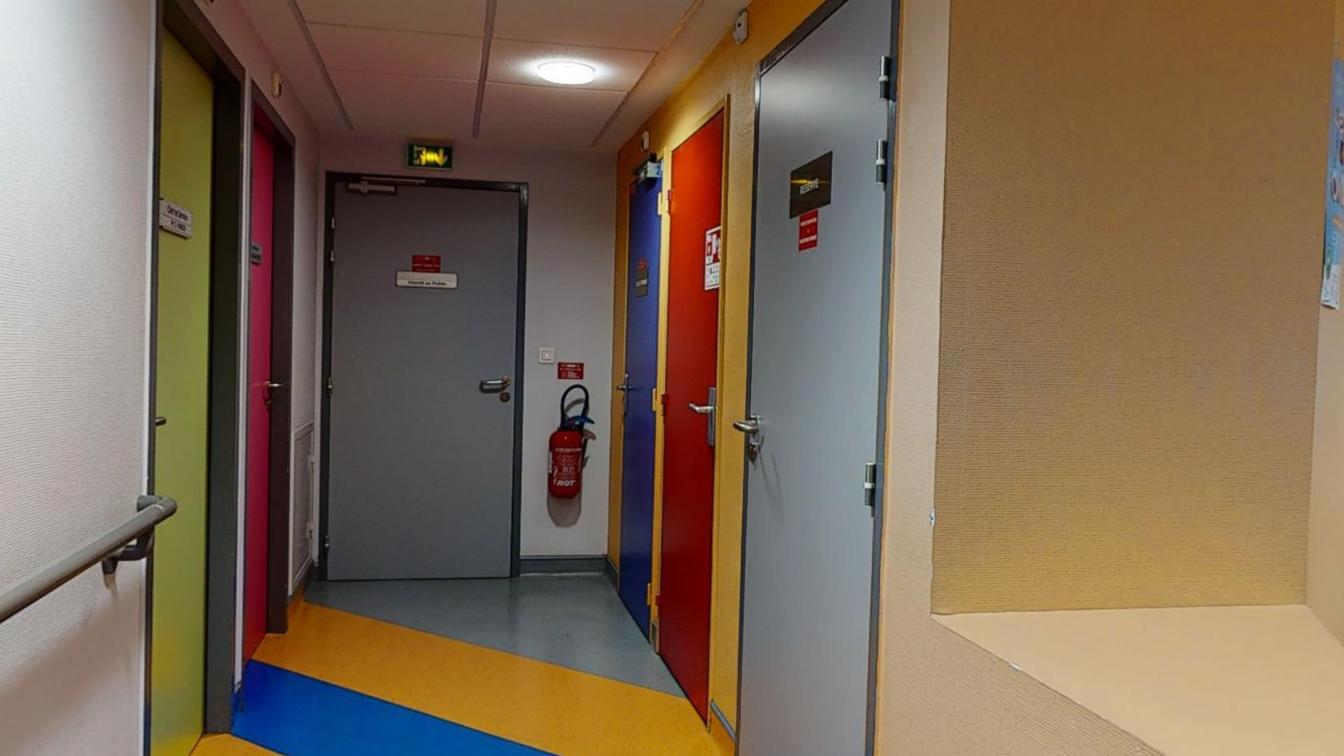}
\includegraphics[width=0.49\textwidth]{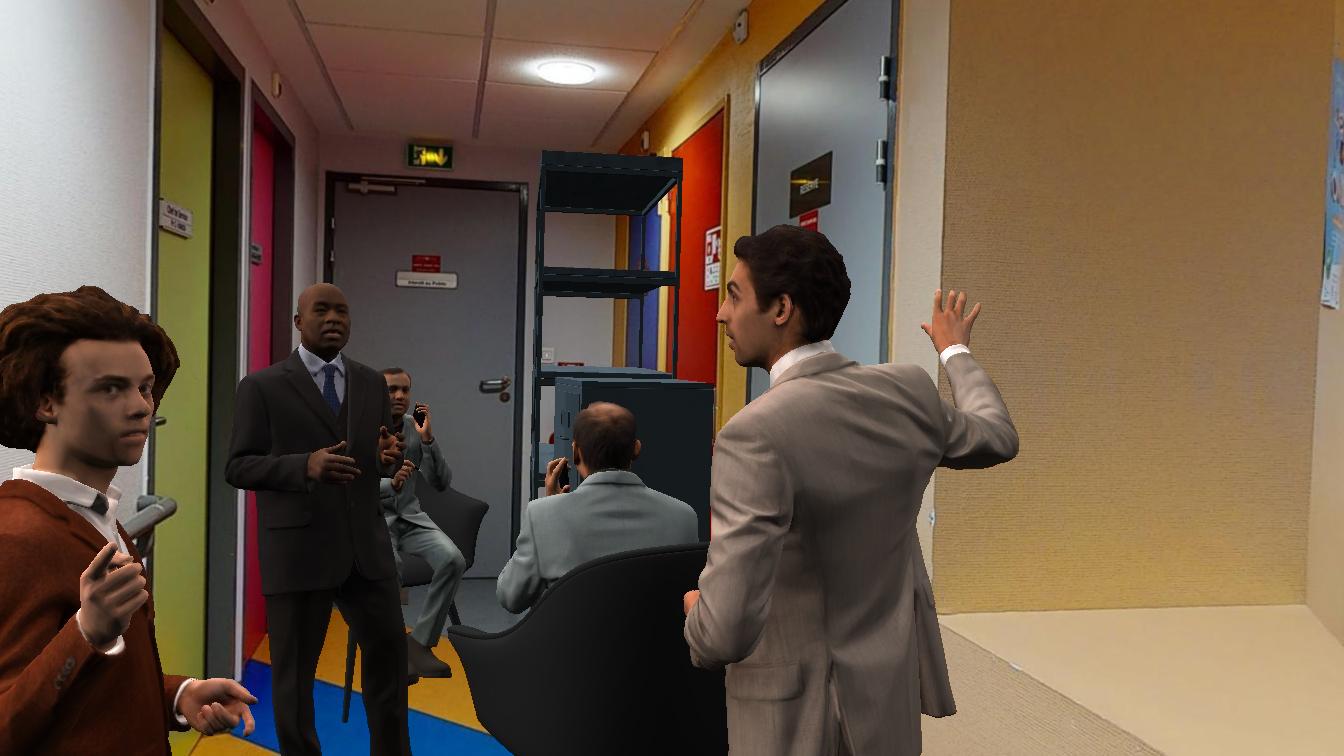}
\includegraphics[width=0.49\textwidth]{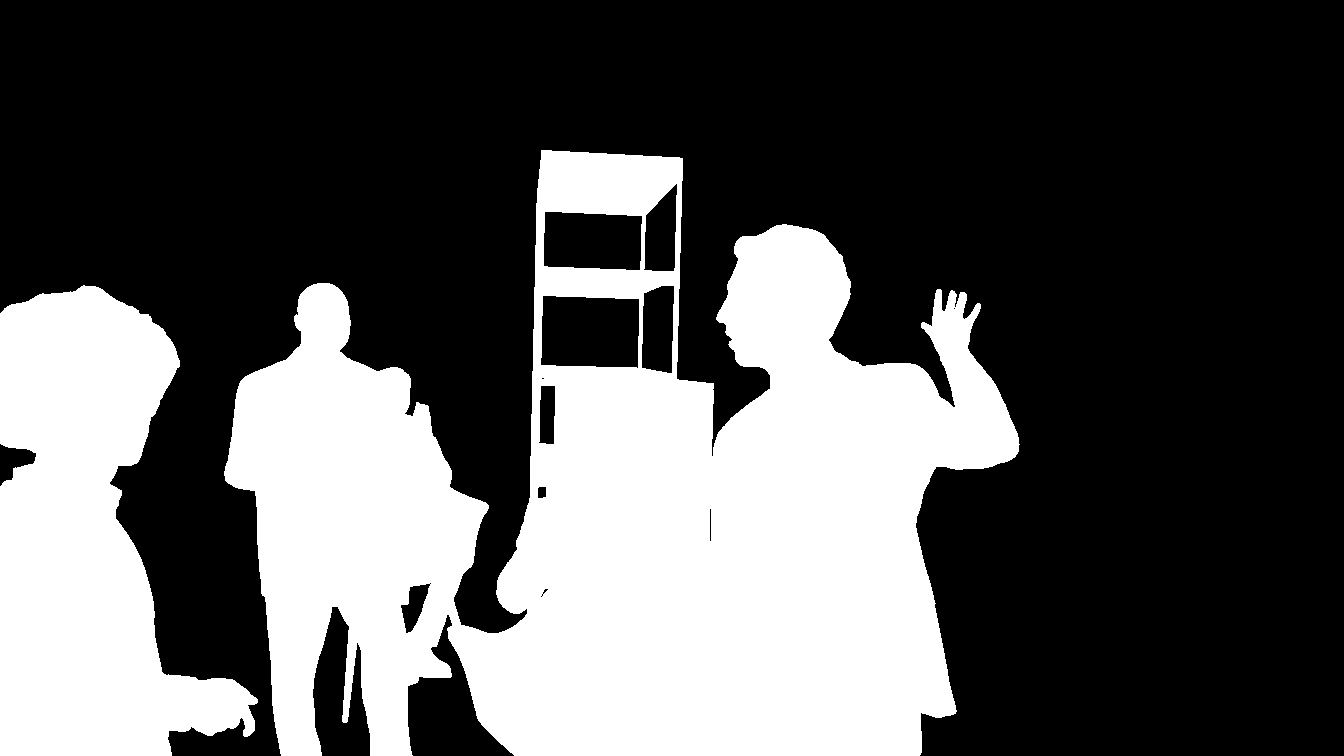} ~
\includegraphics[width=0.49\textwidth]{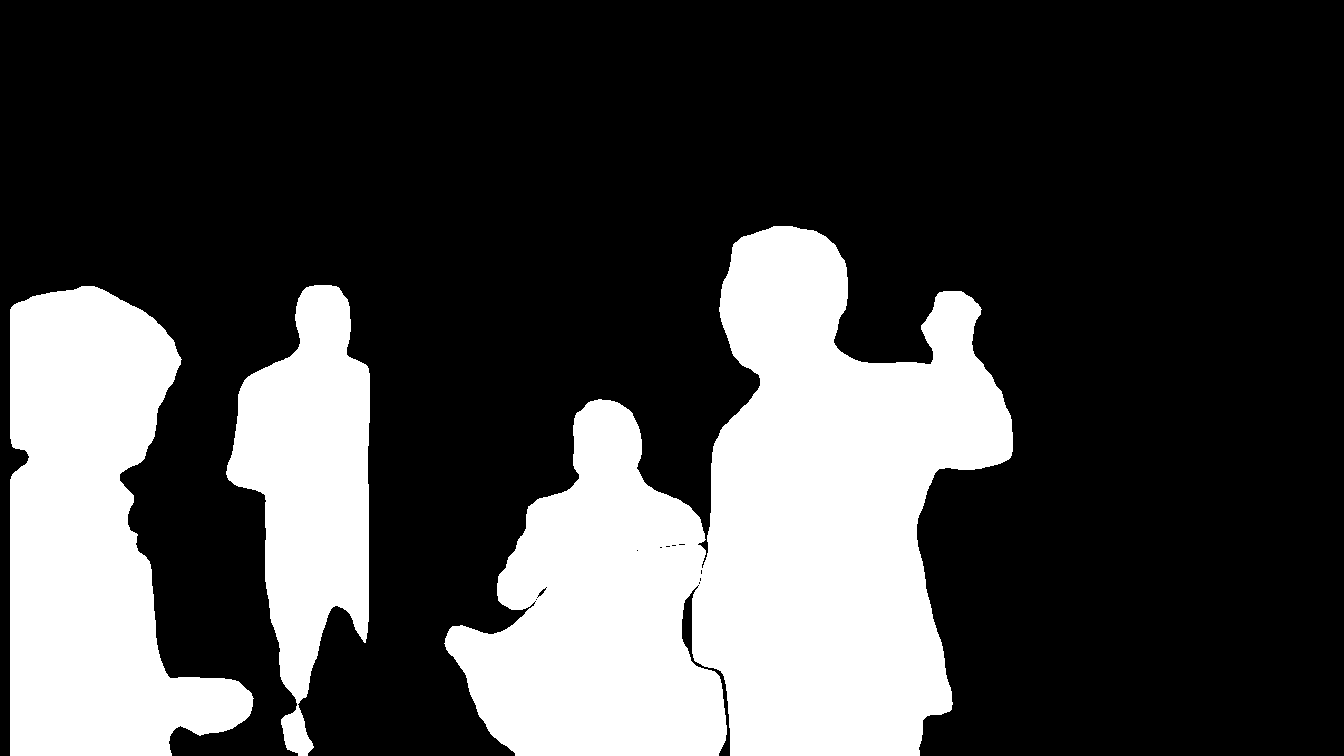}
\caption{The example of the query image and masks. The top left image is a query image without any moving objects. The top right image is a dynamic query with added objects. The bottom left image shows the ground truth masks of the added objects. The bottom right image shows the result of our instance segmentation using YOLACT.} 
\label{fig:dynamic-query}
\end{figure}

The localization pipelines follow common steps, such as relative and absolute pose estimation. These algorithms are, in general, prone to errors. A common issue is that a subset of outliers is marked as inliers. The pose is then optimized to minimize the error \wrt the outliers. Let us show an example of how the semantic masks work on a query image (see Figure~\ref{fig:dynamic-query}).

The relative pose correspondences are filtered according to YOLACT masks (Figure~\ref{fig:dynamic-query}) because of laying on movable objects. It can be seen that a significant part of the correspondences would distort the relative pose estimation step and allow further propagation of mismatches. The image selection step (\ie, the EfficientNet B3 + fully connected layer projection head outputting one number) was trained using 20k image pairs, with $5\%$ employed as test set and the rest as the training set; Adam optimizer with learning rate 0.0001; MSE loss; batch size of 16.

\begin{figure}[H]
\includegraphics[width=1\textwidth]{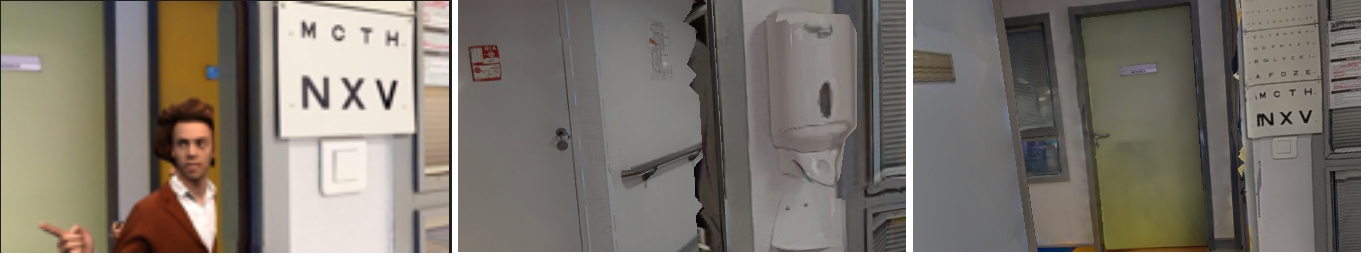}
\vspace{-2em}
\caption{The query image at left and its corresponding rendered synthetic images. The center image shows the best pose candidate using DSIFT only. The right one is using a combination of DSIFT and CNN.} 
\label{fig:cnn_eval}
\end{figure}

\subsubsection{Quantitative Evaluation}
This part shows the performance of the localization with and without filtering moving objects. Our map Dynamic Broca gathers 3636 RGB-D images with known camera poses. We further assume a set of 180 query images. The dataset has 20.63$\%$ of the mean query images occupied by moving objects. The details about the dataset are in Section~\ref{sec:composition-of-test-dataset}.

\begin{figure}[H]
\centering
\includegraphics[width=0.95\textwidth]{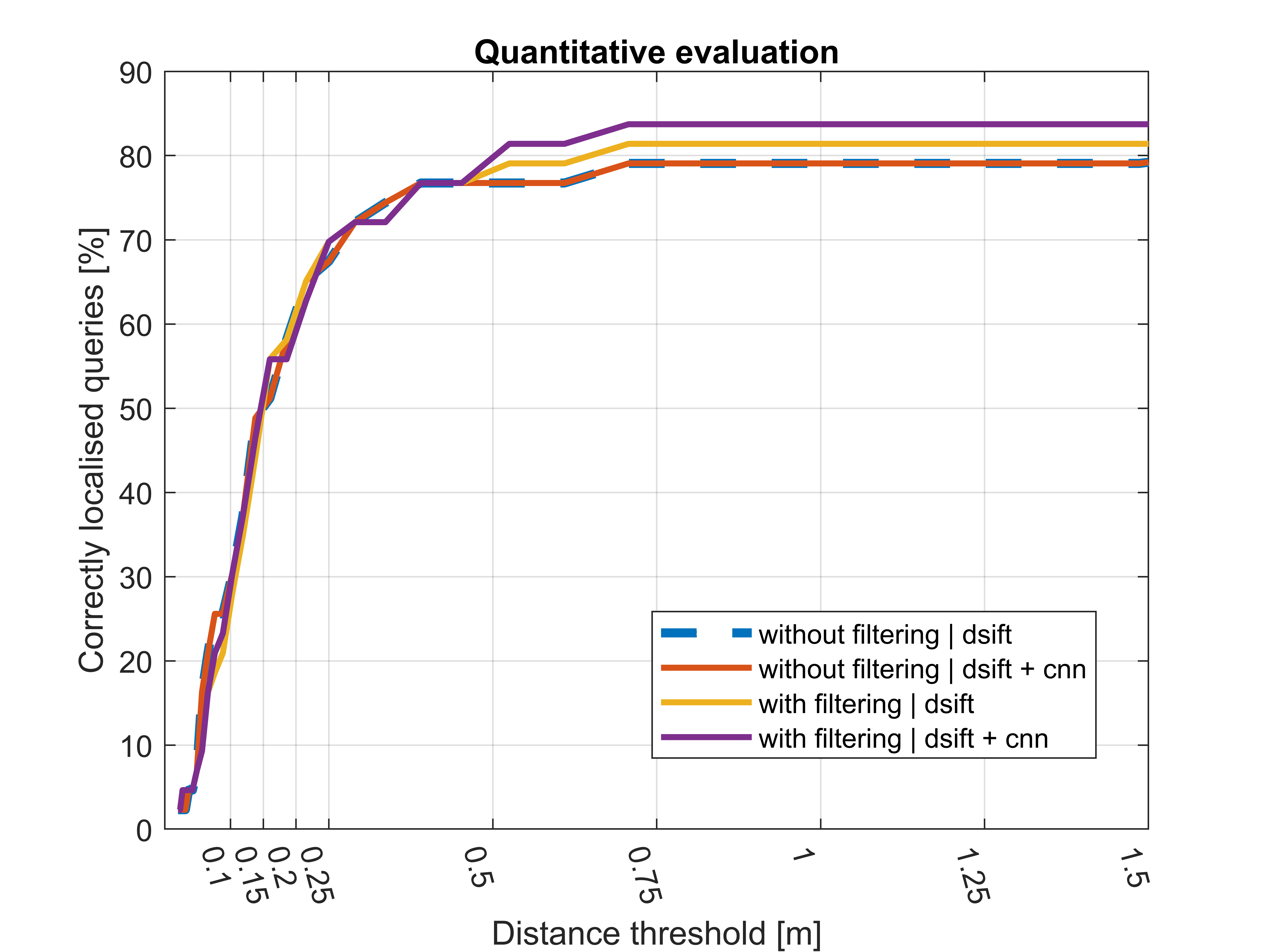}
\vspace{-1em}
\caption{The comparison of original InLoc (blue dashed), InLoc++ (red), D-InLoc (yellow) and D-InLoc++ (purple) on Dynamic Broca dataset. The rotation error threshold is 10$\%$ degrees. We show the localization improvement on images with moving objects occupancy of more than 30$\%$ of the image. The filtering of moving objects (D-InLoc++) improves the overall localization accuracy in comparison with original InLoc algorithm.} \label{fig:quantitative-eval}
\end{figure}

The comparison is shown in Figure~\ref{fig:quantitative-eval}. In our experiments, we assumed YOLACT classes \textit{background}, \textit{TV} and \textit{refrigerator} as static and the class \textit{person} as dynamic object classes. It leads to the $86.9\%$ accuracy of the moving objects masks compared to ground truth masks. The moving object criterion threshold was $\delta = 10^{-9}$. We can see that the proposed straightforward solution for filtering moving objects improves the localization on the dynamic dataset and is further improved with the use of convolutional neural network.

\subsubsection{Conclusion}
This paper opens the essential and not fully explored topic of movable object filtering in the scope of visual localization. We publish a new simple algorithm that detects moving objects based on the query and database images' track lengths and instance segmentation masks. The instance segmentation masks are categorized into moving and static classes and utilized to avoid the propagation of the outliers through the localization process. Mainly, the propagation of the outliers that are close to correct matches and influences the accuracy of camera poses. The results show non-negligible improvement, \ie, we reach approximately 7$\%$ more camera poses within the threshold of 1m for a dataset where movable objects occupy 20.63$\%$ of query images. The second contribution of this paper is that we list the common mistakes of gradient-based image comparison and propose to select the most suitable camera pose by convolutional neural network instead of handcrafted DenseRootSIFT. Lastly, to speed up this area's development, we introduce a new generator of localization datasets with moving objects.

\subsubsection*{Acknowledgements.}
This research was supported by the European Regional Development Fund under IMPACT No.~CZ.02.1.01/0.0/0.0/15 003/0000468, EU H2020 ARtwin No.~856994, and EU H2020 SPRING No.~871245 Projects.

%
%

\clearpage
\bibliographystyle{splncs04}
\bibliography{egbib}

\begin{thebibliography}{10}
\providecommand{\url}[1]{\texttt{#1}}
\providecommand{\urlprefix}{URL }
\providecommand{\doi}[1]{https://doi.org/#1}

\bibitem{cube2sphere}
{cube2sphere 0.2.1}. \url{https://pypi.org/project/cube2sphere/}, accessed:
  2022-5-11

\bibitem{localizationDatasets}
{Localization datasets}. \url{https://www.visuallocalization.net/datasets/},
  accessed: 2022-5-31

\bibitem{matterport}
Matterport. \url{https://matterport.com/}, accessed: 2022-5-20

\bibitem{matterportcamera}
{Mattreport Pro 3D Camera}.
  \url{https://matterport.com/cameras/pro2-3D-camera}, accessed: 2022-5-20

\bibitem{obj2gltf}
{obj2gltf github repository}. \url{https://github.com/CesiumGS/obj2gltf},
  accessed: 2022-5-20

\bibitem{pyrender}
{Pyrender}. \url{https://pyrender.readthedocs.io/en/latest/}, accessed:
  2022-5-14

\bibitem{galileo_accuracy}
Agency, E.S.: Galileo (2021),
  \url{https://gssc.esa.int/navipedia/index.php/Galileo_Performances}

\bibitem{arandjelovic2016netvlad}
Arandjelovic, R., Gronat, P., Torii, A., Pajdla, T., Sivic, J.: Netvlad: Cnn
  architecture for weakly supervised place recognition. In: Proceedings of the
  IEEE conference on computer vision and pattern recognition. pp. 5297--5307
  (2016)

\bibitem{arandjelovic2012three}
Arandjelovi{\'c}, R., Zisserman, A.: Three things everyone should know to
  improve object retrieval. In: 2012 IEEE conference on computer vision and
  pattern recognition. pp. 2911--2918. IEEE (2012)

\bibitem{badue2021self}
Badue, C., Guidolini, R., Carneiro, R.V., Azevedo, P., Cardoso, V.B., Forechi,
  A., Jesus, L., Berriel, R., Paixao, T.M., Mutz, F., et~al.: Self-driving
  cars: A survey. Expert Systems with Applications  \textbf{165},  113816
  (2021)

\bibitem{barath2018graph}
Barath, D., Matas, J.: Graph-cut ransac. In: Proceedings of the IEEE conference
  on computer vision and pattern recognition. pp. 6733--6741 (2018)

\bibitem{barath2020magsac++}
Barath, D., Noskova, J., Ivashechkin, M., Matas, J.: Magsac++, a fast, reliable
  and accurate robust estimator. In: Proceedings of the IEEE/CVF conference on
  computer vision and pattern recognition. pp. 1304--1312 (2020)

\bibitem{bonin2008visual}
Bonin-Font, F., Ortiz, A., Oliver, G.: Visual navigation for mobile robots: A
  survey. Journal of intelligent and robotic systems  \textbf{53}(3),  263--296
  (2008)

\bibitem{chen2021wide}
Chen, K., Snavely, N., Makadia, A.: Wide-baseline relative camera pose
  estimation with directional learning. In: Proceedings of the IEEE/CVF
  Conference on Computer Vision and Pattern Recognition. pp. 3258--3268 (2021)

\bibitem{chum2003locally}
Chum, O., Matas, J., Kittler, J.: Locally optimized ransac. In: Joint Pattern
  Recognition Symposium. pp. 236--243. Springer (2003)

\bibitem{chum2005two}
Chum, O., Werner, T., Matas, J.: Two-view geometry estimation unaffected by a
  dominant plane. In: 2005 IEEE Computer Society Conference on Computer Vision
  and Pattern Recognition (CVPR'05). vol.~1, pp. 772--779. IEEE (2005)

\bibitem{detone2018superpoint}
DeTone, D., Malisiewicz, T., Rabinovich, A.: Superpoint: Self-supervised
  interest point detection and description. In: Proceedings of the IEEE
  conference on computer vision and pattern recognition workshops. pp. 224--236
  (2018)

\bibitem{dusmanu2019d2}
Dusmanu, M., Rocco, I., Pajdla, T., Pollefeys, M., Sivic, J., Torii, A.,
  Sattler, T.: D2-net: A trainable cnn for joint description and detection of
  local features. In: Proceedings of the IEEE/cvf conference on computer vision
  and pattern recognition. pp. 8092--8101 (2019)

\bibitem{localization_benchmark}
Hammarstrand, L., Kahl, F., Maddern, W., Pajdla, T., Pollefeys, M., Sattler,
  T., Sivic, J., Stenborg, E., Toft, C., Torii, A.: Workshop on long-term
  visual localization under changing conditions (2020),
  \url{https://www.visuallocalization.net/workshop/eccv/2020/}

\bibitem{hartley2003multiple}
Hartley, R., Zisserman, A.: Multiple view geometry in computer vision.
  Cambridge university press (2003)

\bibitem{hyeon2020kr}
Hyeon, J., Kim, D., Jang, B., Choi, H., Yi, D.H., Yoo, K., Choi, J., Doh, N.:
  Kr-net: A dependable visual kidnap recovery network for indoor spaces. In:
  2020 IEEE/RSJ International Conference on Intelligent Robots and Systems
  (IROS). pp. 8527--8533. IEEE (2020)

\bibitem{hyeon2021pose}
Hyeon, J., Kim, J., Doh, N.: Pose correction for highly accurate visual
  localization in large-scale indoor spaces. In: Proceedings of the IEEE/CVF
  International Conference on Computer Vision. pp. 15974--15983 (2021)

\bibitem{jin2021image}
Jin, Y., Mishkin, D., Mishchuk, A., Matas, J., Fua, P., Yi, K.M., Trulls, E.:
  Image matching across wide baselines: From paper to practice. International
  Journal of Computer Vision  \textbf{129}(2),  517--547 (2021)

\bibitem{kukelova2008polynomial}
Kukelova, Z., Bujnak, M., Pajdla, T.: Polynomial eigenvalue solutions to the
  5-pt and 6-pt relative pose problems. In: BMVC. vol.~2, p.~2008 (2008)

\bibitem{kukelova2015efficient}
Kukelova, Z., Heller, J., Bujnak, M., Fitzgibbon, A., Pajdla, T.: Efficient
  solution to the epipolar geometry for radially distorted cameras. In:
  Proceedings of the IEEE international conference on computer vision. pp.
  2309--2317 (2015)

\bibitem{kukelova2016efficient}
Kukelova, Z., Heller, J., Fitzgibbon, A.: Efficient intersection of three
  quadrics and applications in computer vision. In: Proceedings of the IEEE
  Conference on Computer Vision and Pattern Recognition. pp. 1799--1808 (2016)

\bibitem{laguna2022key}
Laguna, A.B., Mikolajczyk, K.: Key. net: Keypoint detection by handcrafted and
  learned cnn filters revisited. IEEE Transactions on Pattern Analysis and
  Machine Intelligence  (2022)

\bibitem{larsson2017making}
Larsson, V., Kukelova, Z., Zheng, Y.: Making minimal solvers for absolute pose
  estimation compact and robust. In: Proceedings of the IEEE International
  Conference on Computer Vision. pp. 2316--2324 (2017)

\bibitem{lindenberger2021pixel}
Lindenberger, P., Sarlin, P.E., Larsson, V., Pollefeys, M.: Pixel-perfect
  structure-from-motion with featuremetric refinement. In: Proceedings of the
  IEEE/CVF International Conference on Computer Vision. pp. 5987--5997 (2021)

\bibitem{liu2010sift}
Liu, C., Yuen, J., Torralba, A.: Sift flow: Dense correspondence across scenes
  and its applications. IEEE transactions on pattern analysis and machine
  intelligence  \textbf{33}(5),  978--994 (2010)

\bibitem{lowe2004distinctive}
Lowe, D.G.: Distinctive image features from scale-invariant keypoints.
  International journal of computer vision  \textbf{60}(2),  91--110 (2004)

\bibitem{habitat}
{Manolis Savva*}, {Abhishek Kadian*}, {Oleksandr Maksymets*}, Zhao, Y.,
  Wijmans, E., Jain, B., Straub, J., Liu, J., Koltun, V., Malik, J., Parikh,
  D., Batra, D.: Habitat: {A} {P}latform for {E}mbodied {AI} {R}esearch. In:
  Proceedings of the IEEE/CVF International Conference on Computer Vision
  (ICCV) (2019)

\bibitem{bluetooth_localization}
Marcel, J.: Bluetooth technology is getting precise with positioning systems
  (2019), \url{https://www.bluetooth.com/blog/bluetooth-positioning-systems/}

\bibitem{martin2021nerf}
Martin-Brualla, R., Radwan, N., Sajjadi, M.S., Barron, J.T., Dosovitskiy, A.,
  Duckworth, D.: Nerf in the wild: Neural radiance fields for unconstrained
  photo collections. In: Proceedings of the IEEE/CVF Conference on Computer
  Vision and Pattern Recognition. pp. 7210--7219 (2021)

\bibitem{rpCNN}
Melekhov, I., Ylioinas, J., Kannala, J., Rahtu, E.: Relative camera pose
  estimation using convolutional neural networks. In: International Conference
  on Advanced Concepts for Intelligent Vision Systems. pp. 675--687. Springer
  (2017)

\bibitem{palazzolo2019refusion}
Palazzolo, E., Behley, J., Lottes, P., Giguere, P., Stachniss, C.: Refusion: 3d
  reconstruction in dynamic environments for rgb-d cameras exploiting
  residuals. In: 2019 IEEE/RSJ International Conference on Intelligent Robots
  and Systems (IROS). pp. 7855--7862. IEEE (2019)

\bibitem{Wifi_localization}
Pourhomayoun, M., Fowler, M.: Improving wlan-based indoor mobile positioning
  using sparsity. In: 2012 Conference Record of the Forty Sixth Asilomar
  Conference on Signals, Systems and Computers (ASILOMAR). pp. 1393--1396
  (2012). \doi{10.1109/ACSSC.2012.6489254}

\bibitem{radenovic2018fine}
Radenovi{\'c}, F., Tolias, G., Chum, O.: Fine-tuning cnn image retrieval with
  no human annotation. IEEE transactions on pattern analysis and machine
  intelligence  \textbf{41}(7),  1655--1668 (2018)

\bibitem{revaud2019learning}
Revaud, J., Almaz{\'a}n, J., Rezende, R.S., Souza, C.R.d.: Learning with
  average precision: Training image retrieval with a listwise loss. In:
  Proceedings of the IEEE/CVF International Conference on Computer Vision. pp.
  5107--5116 (2019)

\bibitem{revaud2019r2d2}
Revaud, J., Weinzaepfel, P., De~Souza, C., Pion, N., Csurka, G., Cabon, Y.,
  Humenberger, M.: R2d2: repeatable and reliable detector and descriptor. arXiv
  preprint arXiv:1906.06195  (2019)

\bibitem{rocco2020efficient}
Rocco, I., Arandjelovi{\'c}, R., Sivic, J.: Efficient neighbourhood consensus
  networks via submanifold sparse convolutions. In: European conference on
  computer vision. pp. 605--621. Springer (2020)

\bibitem{runz2018maskfusion}
Runz, M., Buffier, M., Agapito, L.: Maskfusion: Real-time recognition, tracking
  and reconstruction of multiple moving objects. In: 2018 IEEE International
  Symposium on Mixed and Augmented Reality (ISMAR). pp. 10--20. IEEE (2018)

\bibitem{hloc}
Sarlin, P.E., Cadena, C., Siegwart, R., Dymczyk, M.: From coarse to fine:
  Robust hierarchical localization at large scale (2019)

\bibitem{sarlin2020superglue}
Sarlin, P.E., DeTone, D., Malisiewicz, T., Rabinovich, A.: Superglue: Learning
  feature matching with graph neural networks. In: Proceedings of the IEEE/CVF
  conference on computer vision and pattern recognition. pp. 4938--4947 (2020)

\bibitem{sarlin2021back}
Sarlin, P.E., Unagar, A., Larsson, M., Germain, H., Toft, C., Larsson, V.,
  Pollefeys, M., Lepetit, V., Hammarstrand, L., Kahl, F., et~al.: Back to the
  feature: Learning robust camera localization from pixels to pose. In:
  Proceedings of the IEEE/CVF Conference on Computer Vision and Pattern
  Recognition. pp. 3247--3257 (2021)

\bibitem{shavit2019introduction}
Shavit, Y., Ferens, R.: Introduction to camera pose estimation with deep
  learning. arXiv preprint arXiv:1907.05272  (2019)

\bibitem{GPS_accuracy}
for Space-Based Positioning~Navigation, N.C.O., Timing: Gps accuracy (2022),
  \url{https://www.gps.gov/systems/gps/performance/accuracy/}

\bibitem{stenborg2020using}
Stenborg, E., Sattler, T., Hammarstrand, L.: Using image sequences for
  long-term visual localization. In: 2020 International Conference on 3D Vision
  (3DV). pp. 938--948. IEEE (2020)

\bibitem{sun2021loftr}
Sun, J., Shen, Z., Wang, Y., Bao, H., Zhou, X.: Loftr: Detector-free local
  feature matching with transformers. In: Proceedings of the IEEE/CVF
  conference on computer vision and pattern recognition. pp. 8922--8931 (2021)

\bibitem{sun2020acne}
Sun, W., Jiang, W., Trulls, E., Tagliasacchi, A., Yi, K.M.: Acne: Attentive
  context normalization for robust permutation-equivariant learning. In:
  Proceedings of the IEEE/CVF Conference on Computer Vision and Pattern
  Recognition. pp. 11286--11295 (2020)

\bibitem{inloc}
Taira, H., Okutomi, M., Sattler, T., Cimpoi, M., Pollefeys, M., Sivic, J.,
  Pajdla, T., Torii, A.: Inloc: Indoor visual localization with dense matching
  and view synthesis. CoRR  \textbf{abs/1803.10368} (2018),
  \url{http://arxiv.org/abs/1803.10368}

\bibitem{tan2019efficientnet}
Tan, M., Le, Q.: Efficientnet: Rethinking model scaling for convolutional
  neural networks. In: International conference on machine learning. pp.
  6105--6114. PMLR (2019)

\bibitem{tong2021deep}
Tong, W., Matas, J., Barath, D.: Deep magsac++. arXiv preprint arXiv:2111.14093
   (2021)

\bibitem{torii201524}
Torii, A., Arandjelovic, R., Sivic, J., Okutomi, M., Pajdla, T.: 24/7 place
  recognition by view synthesis. In: Proceedings of the IEEE conference on
  computer vision and pattern recognition. pp. 1808--1817 (2015)

\bibitem{torii2013visual}
Torii, A., Sivic, J., Pajdla, T., Okutomi, M.: Visual place recognition with
  repetitive structures. In: Proceedings of the IEEE conference on computer
  vision and pattern recognition. pp. 883--890 (2013)

\bibitem{triggs1999bundle}
Triggs, B., McLauchlan, P.F., Hartley, R.I., Fitzgibbon, A.W.: Bundle
  adjustment—a modern synthesis. In: International workshop on vision
  algorithms. pp. 298--372. Springer (1999)

\bibitem{vidal2020analysis}
Vidal-Balea, A., Blanco-Novoa, O., Picallo-Guembe, I., Celaya-Echarri, M.,
  Fraga-Lamas, P., Lopez-Iturri, P., Azpilicueta, L., Falcone, F.,
  Fern{\'a}ndez-Caram{\'e}s, T.M.: Analysis, design and practical validation of
  an augmented reality teaching system based on microsoft hololens 2 and edge
  computing. In: Engineering Proceedings. vol.~2, p.~52. Multidisciplinary
  Digital Publishing Institute (2020)

\bibitem{yi2018learning}
Yi, K.M., Trulls, E., Ono, Y., Lepetit, V., Salzmann, M., Fua, P.: Learning to
  find good correspondences. In: Proceedings of the IEEE conference on computer
  vision and pattern recognition. pp. 2666--2674 (2018)

\bibitem{zhou2021patch2pix}
Zhou, Q., Sattler, T., Leal-Taixe, L.: Patch2pix: Epipolar-guided pixel-level
  correspondences. In: Proceedings of the IEEE/CVF conference on computer
  vision and pattern recognition. pp. 4669--4678 (2021)

\end{thebibliography}

\end{document}